# A Multiple-Instance Learning Approach for the Assessment of Gallbladder Vascularity from Laparoscopic Images

Constantinos Loukas
*Laboratory of Medical Physics,
Medical School
National and Kapodistrian University
of Athens*
Athens, Greece
cloukas@med.uoa.gr

Athanasios Gazis
*Laboratory of Medical Physics,
Medical School
National and Kapodistrian University
of Athens*
Athens, Greece
thgazis@med.uoa.gr

Dimitrios Schizas
*1st Department of Surgery, Laikon
General Hospital
National and Kapodistrian University
of Athens*
Athens, Greece
schizasad@gmail.com

*Abstract*—An important task at the onset of a laparoscopic cholecystectomy (LC) operation is the inspection of gallbladder (GB) to evaluate the thickness of its wall, presence of inflammation and extent of fat. Difficulty in visualization of the GB wall vessels may be due to the previous factors, potentially as a result of chronic inflammation or other diseases. In this paper we propose a multiple-instance learning (MIL) technique for assessment of the GB wall vascularity via computer-vision analysis of images from LC operations. The bags correspond to a labeled (low *vs.* high) vascularity dataset of 181 GB images, from 53 operations. The instances correspond to unlabeled patches extracted from these images. Each patch is represented by a vector with color, texture and statistical features. We compare various state-of-the-art MIL and single-instance learning approaches, as well as a proposed MIL technique based on variational Bayesian inference. The methods were compared for two experimental tasks: image-based and video-based (i.e. patient-based) classification. The proposed approach presents the best performance with accuracy 92.1% and 90.3% for the first and second task, respectively. A significant advantage of the proposed technique is that it does not require the time-consuming task of manual labelling the instances.

*Keywords—surgery, laparoscopic cholecystectomy, gallbladder, vascularity, classification, multiple-instance learning.*

## I. INTRODUCTION

Laparoscopic Surgery (LS), offers substantial benefits for the patient such as minimized blood loss, rapid recovery, better cosmetic results and lower risk of infection. In addition, the laparoscopic camera allows to record the video of the surgery, thus providing a rich set of visual information that can be leveraged for various computer vision applications. For example, vision-based systems may provide context-aware assistance to the surgeon during or after the operation to facilitate improvements in the delivery of surgical care. Artificial Intelligence (AI) in surgery has a key role in this direction by training a computer to analyze and understand images and ultimately enhance surgical performance throughout the patient care pathway [1].

To date, surgical video analysis has been employed to provide key semantic information about the status of an operation, such as its current phase [2], remaining duration [3], instrument detection [4], and coagulation events [5]. Post-operatively, the video recordings have been employed for surgical performance analysis [6], keyframe extraction [7], surgical gesture recognition [8], and management of large-scale surgical data repositories [9]. In addition, the availability of annotated video datasets, such as Cholec80 [10], has been a key factor in recent AI applications, allowing the employment of state-of-the-art machine learning techniques such as deep learning.

Apart from recording the video of the operation, the surgeon may also acquire still frames that reflect certain visual features of the operated organ or the outcome of a procedural task. Moreover, the acquired images may be utilized post-operatively in the patient's formal report, for future reference about the patient's anatomy and for medical education or research purposes [11]. Recently, a few research works have been published on the analysis of still frames extracted from the video of a surgery. A framework for multi-label classification of laparoscopic images into five anatomic classes was proposed in [12]. In [13] a convolutional neural network (CNN) was employed for semantic image segmentation into surgical tools and gynecologic anatomic structures. In [14] a computer vision technique was applied to surgical images with the aim to assess surgical performance on pelvic lympl node dissection. In [15], visual features extracted from manual annotations of still video frames were employed to assess surgical exposure, an important indicator of surgical expertise. AI on surgical images has also been applied recently for intraoperative guidance and detection of adverse events [16],[17].

In this paper we elaborate from a different perspective on the image-based assessment of gallbladder (GB) vascularity proposed recently in [18]. In laparoscopic cholecystectomy (LC), a common surgical technique for the treatment of GB diseases, the surgeon initially inspects the GB to assess certain features that are important for the strategy to be performed. Some of these features include thickness of the GB wall, indications of inflammation and fat coverage. In addition, difficulty in the visualization of the GB vessels may result from fatty infiltration or increased thickening of the GB wall (potentially as a result of chronic inflammation or other diseases), conditions that may suggest increased intraoperative difficulty [19].

In [18] the main focus was on the vascularity assessment of GB images extracted from LC videos. The employed method was based on a CNN trained to classify the GB wall into vascularity levels (e.g. low *vs.* high). Due to the small number of GB images available for CNN training, 800 image patches were manually annotated by surgical experts. Although highly promising (91.7% accuracy), this approach was based on the time consuming task of patch annotation to create the ground truth dataset.

In the present work we propose an alternative strategy that alleviates this limitation. In particular, we employ a different machine learning approach based on multiple-instance learning (MIL). Compared to single-instance learning (SIL), in MIL the training examples correspond to bags of instances and each bag is assigned a specific label. In the present work, the bags correspond to the limited dataset of GB images available, and the instances correspond to sequential patches extracted from the GB images. It is important to note that the MIL approach does not require that the labels of the instances are known, but rather the labels of the bags. This is particularly suited in our case since the number of GB images available is limited, whereas from each GB image we can extract dozens of patches, without the need to annotate them.

The rest of the paper is organized as follows. The next section presents the employed dataset, the image feature extraction process and various state-of-the-art MIL techniques employed, as well as a proposed one based on variational Bayesian inference. Subsequently follow the comparative experiments and presentation of the results for two experimental tasks (image-based and video-based classification). Finally, we conclude the paper with a discussion and directions for future work.

## II. METHODOLOGY

### A. Dataset

For the purpose of this study we employed the GBVasc181 dataset[1] [18], which includes 181 surgical images with manual contours (regions of interest-ROIs) of the GB wall. The images are extracted from 53 LC videos of the Cholec80 dataset [10] and the ROIs contain the body and fundus of the GB. In addition, the dataset provides labels with respect to the vascularity level of the GB ROIs: low ($L$) and high ($H$) vascularity. $H$ denotes presence of prominent superficial vessels whereas $L$ denotes absence of vessels or extensive fat coverage. The annotation was performed by two expert surgeons (E1 and E2) via visual inspection of the GB ROIs. According to [18], the level of agreement between the two experts was high (~92%), so we randomly chose the annotations of expert E1 as the ground-truth for algorithm training.

TABLE I. DATA STATISTICS. NUMBER (#) OF GB IMAGES AND PATCHES PER VASCULARITY CLASS ($L,H$).

| Vascularity class: | | $L$ | $H$ |
|---|---|---|---|
| # annotated GB images (ROIs): | | 71 | 110 |
| # patches extracted from the GB ROIs: | Min | 2 | 1 |
| | Max | 51 | 70 |
| | Median | 14 | 17 |
| | Total | 1214 | 2058 |

The GBVasc181 dataset also includes vascularity annotation for 800 image patches extracted selectively from the ROIs, but this information was not used in this study. Instead, each GB ROI was considered as a bag of instances corresponding to image patches (64x64) extracted sequentially in a sliding window fashion with 50% overlap. In total 3,272 patches were extracted from the 181 ROIs. Table 1 provides a statistical overview of the employed ROI/patch dataset and the ground-truth annotations.

[1] https://mpl-en.med.uoa.gr/as/datasets

It should be noted that the patch labels were considered unknown. Indeed, the patches extracted from a GB ROI may not necessarily inherit the label of the ROI. A GB ROI with vascularity $H$ may also contain patches from the $L$ class, and vice versa. As shown in Fig. 1 the GB wall presents a variable vascularity pattern with regions from both classes. Hence, the employed dataset is in essence weakly labeled and the MIL paradigm is particularly suited: only the labels of the bags (ROIs) is known, whereas the label of the instances (patches) contained in each bag is unknown.

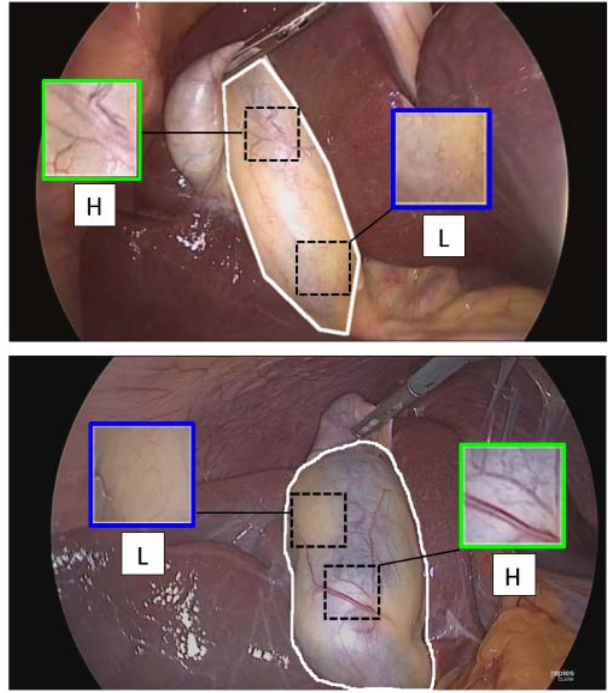

Fig. 1. Examples of GB images from the GBVasc181 dataset. The manual outline of the ROI is shown in white. The ground-truth vascularity label for the top and bottom ROI is $L$ and $H$, respectively. Insets show sample patches with different vascularity.

### B. Feature Extraction

Each patch was represented by a 707-long feature vector that included 3 categories of features [20]. First, color and color-edge features were extracted after the patches were quantized to 32 colors via k-means. Compared to uniform color quantization, the previous approach was preferred since the color values expanded over a limited region in the RGB space. The number of color features was 259 and included: mean RGB color values, color histogram, auto color correlogram, improved color coherence, and color edge magnitude-direction histograms. Color coherence considers the size and locations of the regions with a particular quantized color and the auto color correlogram measures how often a quantized color finds itself in its immediate neighborhood. The color edge magnitude-direction histograms employs the Sobel gradient operator on each quantized color image plane providing two histograms for the edge magnitude and direction.

The second category included mostly information about texture (405 features). The input was the intensity component of the color image and the extracted features were: histogram of oriented gradients (HOG) with 7 bins and cell-size=16, Tamura features (coarseness, contrast and directionality), and the edge histogram descriptor.

The third category included 43 statistical features extracted from the RGB image. In particular, we extracted global features such as skewness and kurtosis as well as statistical features using the following higher-order matrix-based types: GLCM (gray-level co-occurrence matrix), GLRLM (gray-level run-length matrix), GLSZM (gray-level size zone matrix) and NGTDM (neighborhood gray-tone difference matrix). From each matrix various statistical features were extracted such as: energy, contrast, etc. (GLCM); short run emphasis, long run emphasis, etc. (GLRLM); small zone emphasis, large zone emphasis, etc. (GLSZM); complexity, strength, etc. (NGTDM). It should be noted that compared to the standard calculation of these matrices based on 8-heignboors connectivity, in our case the RGB image was considered a 3D volume. Thus, the matrices were obtained using 26-neighboors connectivity (i.e. pixels were considered to be neighbors in all 13 directions in the three dimensions).

After feature extraction, all features were concatenated into a patch-based feature vector with 707 dimensions. To reduce the dimensionality of the feature space, PCA was performed using the feature vectors collected from all patches of the training set of GB images. This process led to feature vectors with 211-223 dimensions (depending on the training fold), that accounted for > 95% of the total variance.

*C. MIL Methods*

MIL is a form of supervised learning applicable to problems where the training examples correspond to bags of instances and each bag is assigned a specific label but the labels of the instances are usually unknown. Under the MIL formulation, $\{(\boldsymbol{X}_i, L_i)\}_{i=1}^{N}$ is a data set of $N$ training bags $\boldsymbol{X}_i$ and each bag is associated with label $L_i$. A bag contains a set of $m_i$ instances: $\boldsymbol{X}_i = \{\boldsymbol{x}_{ij}\}_{j=1}^{m_i}$, usually in form of a feature vector, whereas the number of instances $m_i$ may vary among the bags.

The MIL formulation was first proposed to solve the binary *musk* drug activity prediction problem [21]: a molecule (bag of instances) is considered active (resp. inactive) if one (resp. none) of its spatial confirmations (instances) is able to bind to a certain target site. The solution to this problem was approached via the standard MIL assumption, which states that a positive bag contains at least one positive instance, whereas negative bags contain only negative instances:

$$L = \begin{cases} +1, \exists\ l_j: l_j = +1 \\ -1, \forall\ l_j: l_j = -1 \end{cases} \quad (1)$$

where $l_j = \{+1, -1\}$ denotes the hidden class label for an instance $\boldsymbol{x}_j$ that belongs to bag $\boldsymbol{X}$.

MIL algorithms may be categorized according to their ability to perform instance-level or bag-level predictions [22]. The first category (primarily instance-level) targets instance prediction but may easily be employed for bag prediction, as required for the purpose of this study. After training an instance-level detector, a bag is positive if it contains at least one positive instance otherwise it is negative. This category includes algorithms such as the Axis-parallel hyper rectangle (APR), Diversity Density (DD), its variant Expectation-Maximization DD (EM-DD), and mi-SVM. The aforementioned approaches follow the standard MIL assumption.

The second category includes bag-level algorithms that are optimized for bag-level prediction and can be further divided according to their ability to also perform instance-level prediction (primary bag-level) or not (exclusively bag-level). A well-known primary bag-level algorithm is MI-SVM and it follows the standard MIL assumption. From the exclusively bag-level subcategory we employed the Citation KNN (CKNN) and mi-Graph, both of which follow alternative MI assumptions (nearest neighbor and graph assumptions, respectively). For an extensive review of MIL techniques the reader is referred to [22],[23]. In the following we summarize the aforementioned MIL algorithms and a proposed method (MI-VBGMM) that falls under the exclusively bag-level category. All methods were employed for the GB vascularity classification problem studied in this paper.

The APR [21] was the first method that introduced the MIL paradigm, aiming to find a hyper rectangle that contain at least one instance from each positive bag while excluding all the instances from negative bags. A bag is classified as positive (resp. negative) if one (resp. none) of its instances lies within the APR.

For the DD method [24], the goal is to find a discriminative point in the feature space so that in its neighborhood all positive bags have at least on instance, while instances from negative bags are far away. The location of this point and the feature weights defining the appropriate neighborhood are found by maximizing a DD metric. The EM-DD [25] is a widely known variant that employs the EM algorithm.

The mi-SVM and MI-SVM [26] both employ the maximum margin concept of the SVM algorithm. The mi-SVM method focuses on instance-level prediction by maximizing the separation between positive and negative instances. The goal of MI-SVM is to maximize positive and negative bags by focusing on the most 'most positive' and 'least negative' instances contained in the positive and negative bags, respectively.

CKNN generalizes the k-nearest neighbors (k-NN) idea using the Hausdorff distance as the bag-level distance metric [27]. In addition to the nearest neighbors of a bag, citers that count the candidate bag as one of their neighbors are also considered in the classification rule.

The mi-Graph relies on the assumption that the spatial relationships among instances are important for the label of the bag. This method employs SVM classification at the bag-level using a kernel that is based on the ε-graph representation of the bags [28].

The proposed method (MI-VBGMM) belongs to the exclusively bag-level subcategory and employs variational Bayesian Gaussian mixture models (VBGMM) at its core. The overall process consists of 3 main steps. First, the instances from all bags of the training set are clustered using VBGMM. Compared to other techniques (k-means, k-medoids, GMM, etc.), VBGMM does not require to predefine the number of clusters. Starting with an initial (usually high) number of clusters $K$, the optimum number of clusters $K^*$ is estimated by excluding components with small weights (<1%). The weights are computed via a formula that provides the expected value of the mixing coefficients involved in the GMM. In this study $K^*$ was found about 24 (depending on the training fold). A detailed implementation of the algorithm with application to surgical images is provided in [12,29]. Second, each instance $x_{ij}$ is represented by the probability $\gamma_{ijk}$ that the instance is assigned to each of the $K$ clusters. The formula for $\gamma_{ijk}$ is omitted for brevity. Finally, each bag $X_i$ is represented by a novel feature vector $z_i$ that accumulates $\gamma_{ijk}$ of all $m_i$ instances for the $K^*$ components:

$$z_i = c \times \left[ \sum_{j=1}^{m_i} \gamma_{ij1}, \ldots, \sum_{j=1}^{m_i} \gamma_{ijK^*} \right] \quad (2)$$

where $c$ is a normalization factor so that $\|z_i\|_1 = 1$.

Having obtained the VBGMM parameters, the previous step is also employed to transform the test bags into novel feature vectors $z_i$. Finally, after having a vector representation for every bag, the MIL problem is transformed into a standard classification task which was addressed via the SVM algorithm. After SVM training, a test bag is classified as positive (resp. negative), if the corresponding SVM score is positive (resp. negative).

### D. Experimental Protocol

The 53 videos were randomly split into five-folds, so that every fold included GB images (and their corresponding patches), from different operations (i.e. patients). Based on a five-fold cross-validation, one of the five folds served as the test set (20%) and the other four folds as the training set (80%).

We evaluated two experimental tasks. For the first task (*image level* classification) each image is considered as a bag, annotated with its vascularity label (*L* or *H*), and the instances are the patches extracted from the images. On average there are 3.4 images per video and 18 patches per image (i.e. a single bag contains about 18 instances). For each experimental run the training and test sets included approximately 145 and 36 bags, respectively.

For the second task (*video level* classification), we evaluated the algorithm's performance at the video level based on majority voting of the image labels predicted from the first task. In particular, the label with most votes was assigned to the video. In case the label counts of the two classes were equal, the label with the highest probability was considered. For the ground truth video labels we did not encountered such an issue. For every video the majority of its image labels (annotated by the expert surgeon) corresponded to a single class.

## III. RESULTS

For both experimental tasks the performance of each algorithm was evaluated via the metrics: accuracy (Acc), precision (Pre), recall (Rec), and F1. The results are presented as mean values across the five experimental runs. The convention used was: the label "positive" (resp. "negative") refers to *H* (resp. *L*) vascularity images.

Eleven methods, as described in Section II.C, were evaluated on the GBVasc181 dataset: APR, CKNN, DD, EM-DD, mi-Graph, mi-SVM and MI-SVM and MI-VBGMM (the last three with linear and RBF kernels for the SVM classifier). For all methods, except mi-Graph and MI-VBGMM, we used the implementation reported in [30]. The mi-Graph was obtained from [28] and the reference website of the authors (LAMDA). The proposed method MI-VBGMM was implemented based on [29] and our previous work [12]. For each method the hyper-parameters were optimized using grid search. In the following we first present the result for the two experimental tasks and then we assess the best MIL method against various SIL methods.

### A. Comparison of MIL Methods

In the first experimental task (image-level classification), the goal was to predict the vascularity label of the GB ROIs using the extracted patches. Table 2 shows the performance of the examined methods. As expected, APR, DD and EM-DD have the lowest performance, probably because positive instances do not form a single cluster in the feature space. The mi-SVM approach has the highest performance among the primarily instance-level algorithms (87.5% Acc). However, the bag-level algorithms (mi-Graph, MI-SVM, CKNN and MI-VBGMM) outperform all instance-level algorithms (88.2-92.1% Acc). CKNN and MI-SVM have similar performance (88.2% and 88.9% Acc), whereas mi-Graph shows slightly better performance (90.2% Acc). For the SVM-based methods, the linear kernel seems to be slightly more suitable compared to the RBF kernel. The proposed method shows the best performance across most metrics: 92.1% Acc, 94.6% Pre and 94.0% F1. As reported in [18], the agreement at the image-level between two expert surgeons was close to 92%, which is similar to the performance of MI-VBGMM linear.

TABLE II. MIL PERFORMANCE COMPARISON FOR IMAGE-LEVEL CLASSIFICATION. BEST RESULTS COLUMN-WISE ARE IN BOLD.

| Method | Acc (%) | Pre (%) | Rec (%) | F1 (%) |
|---|---|---|---|---|
| APR | 65.2 | 68.0 | 87.2 | 76.4 |
| CKNN | 88.2 | 88.3 | 93.8 | 91.0 |
| DD | 41.3 | 93.3 | 9.7 | 17.6 |
| EM-DD | 66.9 | 67.1 | 95.0 | 78.6 |
| mi-Graph | 90.2 | 93 | 91.8 | 92.4 |
| mi-SVM linear | 87.5 | 85.1 | 97.4 | 90.8 |
| mi-SVM RBF | 86.9 | 82.9 | **100.0** | 90.7 |
| MI-SVM linear | 88.9 | 92.2 | 90.4 | 91.3 |
| MI-SVM RBF | 88.5 | 93.0 | 88.9 | 90.9 |
| MI-VBGMM linear (proposed) | **92.1** | **94.6** | 93.4 | **94.0** |
| MI-VBGMM RBF (proposed) | 91.1 | 93.8 | 92.4 | 93.1 |

The next goal was to predict the vascularity label of the patient's GB using the images extracted from the video of the operation (video-level classification). Table 3 shows the results for this experimental task. As described before, a majority-voting approach was employed using the image labels predicted from the first task. Similarly to the previous results, APR, DD and EM-DD have the lowest performance and mi-SVM shows the highest performance (85.3% Acc) among the four instance-level methods. The accuracy of the four bag-level algorithms (mi-Graph, MI-SVM, CKNN and MI-VBGMM) is again higher than that of the instance-level ones (86.9-90.3% Acc). The proposed method presents the best performance, higher than 90%, across all metrics: 90.3% Acc, 93.8% Pre, 92.1% Rec and 92.9% F1. For all SVM-based methods the linear kernel results in a slightly better performance. Moreover, the accuracy of the proposed method is the highest than all other methods, independent to the SVM kernel employed.

TABLE III. MIL PERFORMANCE COMPARISON FOR VIDEO-LEVEL CLASSIFICATION. BEST RESULTS COLUMN-WISE ARE IN BOLD.

| Method | Acc (%) | Pre (%) | Rec (%) | F1 (%) |
|---|---|---|---|---|
| APR | 68.2 | 71.6 | 90.3 | 79.9 |
| CKNN | 86.9 | 87.7 | 94.6 | 91.0 |
| DD | 39.8 | 93.3 | 14.6 | 25.2 |
| EM-DD | 68.6 | 70.4 | 94.6 | 80.7 |
| mi-Graph | 88.6 | 91.5 | 92.1 | 91.8 |
| mi-SVM linear | 85.3 | 84.3 | 96.8 | 90.1 |
| mi-SVM RBF | 85.8 | 83.1 | **100.0** | 90.8 |
| MI-SVM linear | 88.1 | 91.4 | 91.5 | 91.4 |
| MI-SVM RBF | 87.5 | 92.0 | 90.0 | 91.0 |
| MI-VBGMM linear (proposed) | **90.3** | **93.8** | 92.1 | **92.9** |
| MI-VBGMM RBF (proposed) | 89.7 | 92.9 | 92.3 | 92.6 |

*B. MIL vs. SIL Methods*

Based on the same setup employed for the MIL experiments (i.e. experimental tasks and training/test folds), we compared the proposed method (MI-VBGMM) against five SIL methods: SVM (with linear kernel), k-nearest neighbors (kNN), naïve Bayes (NB), random forest (RF), and AdaBoost. Hyper-parameter optimization was again performed via grid search. In contrast to MIL, SIL methods consider as samples the instances from all bags. For each instance, the ground-truth label was assigned to that of the bag it belongs to. After training, the label of a candidate bag is predicted via majority voting of the predicted labels of its instances. Hence, for the first task (image-level classification) the label of a GB ROI is determined by majority voting of the patch predicted labels. For the second task (video-level classification), the label prediction approach was the same to that followed for the MIL methods, as described previously.

Tables 4 and 5 show the results for the first and second experimental tasks, respectively. The proposed MIL method outperforms all other SIL methods across most metrics. In particular, for image-level classification the accuracy and F1 metric of MI-VBGMM is higher by 3.6% and 2.5% compared to the second best method, respectively (92.1% *vs.* 88.5%-SVM and 94.0% *vs.* 91.5%-AdaBoost). Moreover, MI-VBGMM outperforms the CNN-based method reported in [18]. In terms of accuracy, the performance is slightly higher (92.1% *vs.* 91.2%) whereas for the other metrics the performance difference is notably higher (4.2% for Pre, 1.9% for Rec and 3.1% for F1). Note that the single instance CNN method in [18] employs 800 manually labelled patches for CNN training. In contrast, MI-VBGMM training is based only on manual labelling of the GB ROIs (181 images; same dataset as in [18]), resulting in a significant reduction in the annotation cost.

For the video-level classification (Table 5), the proposed method presents again the best performance. The accuracy and F1 metric of MI-VBGMM is higher by 2.5% and 0.8% compared to the second best method (AdaBoost), respectively (90.3% *vs.* 87.8% and 92.9% *vs.* 92.1%).

Fig. 2 shows the normalized confusion matrices for the best method (MI-VBGMM linear), for the image- and video-level classification tasks. The normalization was applied on the aggregation of the confusion matrices across the five test-sets. It may be seen that the *H* class is recognized better than the *L* class in both experimental tasks (93.4 *vs.* 89.9 and 92.2 *vs.* 85.7). This may be due to the fact that the presence of blood vessels in the *H* class images provide a distinguishable pattern that is captured more easily by the extracted features. In contrast, *L* images are distinguishable only by their color (yellowish due to great fat coverage), and they lack a texture pattern due to the absence of blood vessels.

TABLE IV. COMPARISON OF MIL *VS.* SIL METHODS FOR IMAGE-LEVEL CLASSIFICATION.

| Method | Acc (%) | Pre (%) | Rec (%) | F1 (%) |
|---|---|---|---|---|
| MI-VBGMM linear (proposed) | **92.1** | **94.6** | 93.4 | **94.0** |
| SVM | 88.5 | 88.9 | 91.4 | 90.1 |
| kNN | 86.9 | 88.6 | 92.9 | 90.7 |
| NB | 80.3 | 77.8 | 96.9 | 86.3 |
| RF | 87.2 | 86.3 | 96.4 | 91.1 |
| AdaBoost | 86.9 | 86.0 | **97.7** | 91.5 |
| CNN | 91.2 | 90.4 | 91.5 | 90.9 |

TABLE V. COMPARISON OF MIL *VS.* SIL METHODS FOR VIDEO-LEVEL CLASSIFICATION.

| Method | Acc (%) | Pre (%) | Rec (%) | F1 (%) |
|---|---|---|---|---|
| MI-VBGMM linear (proposed) | **90.3** | **93.8** | 92.1 | **92.9** |
| SVM | 80.0 | 84.0 | 87.5 | 85.7 |
| kNN | 81.1 | 83.3 | 92.6 | 87.7 |
| NB | 78.1 | 78.0 | 96.1 | 86.1 |
| RF | 86.4 | 87.1 | 95.3 | 91.0 |
| AdaBoost | 87.8 | 87.9 | **96.7** | 92.1 |

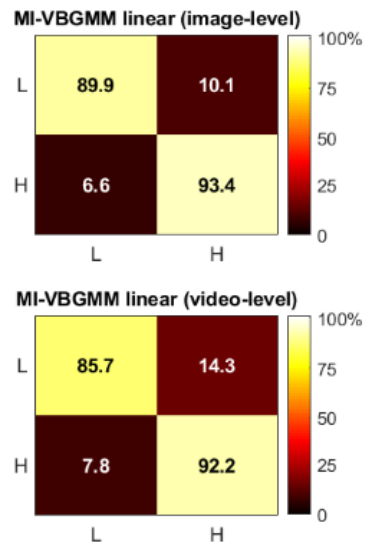

Fig. 2. Color-coded confusion matrices of the proposed method for the two experimental tasks (image-level and video-level classification). The X and Y-axis represent predicted and ground truth labels, respectively.

IV. CONCLUSIONS

In this paper we investigate the potential of image-based assessment of the GB wall vascularity from intraoperative images using various state-of-the art MIL techniques and a proposed bag-level approach based on VBGMMs and SVM. In addition, we compared the best MIL method with SIL techniques in order to assess the significance of the multiple instance concept over standard classification of single instances.

Our results show that the MIL framework is particularly suited to the problem of GB vascularity classification. The GB image can be considered as a bag, labelled with its vascularity level, whereas the instances are patches extracted from the GB ROI. For the image-level classification task, the proposed approach presents the best performance with Acc 92.1%. For the video-level classification task the accuracy was slightly lower, 90.3%. In terms of the SIL methods, the best approach was based on a CNN and provided slightly lower performance (91.2% Acc for image-level classification). However, CNN training requires manual annotation for a large number of patches, which is tedious and time-consuming. In contrast, the proposed MIL technique requires only manual annotation of the GB images, the number of which is significantly lower. MIL can thus leverage surgical image classification to improve GB vascularity assessment, without the need of labeling the patches extracted from every surgical image.

A potential extension of the proposed work is to apply the MIL concept directly at the patient-level. In particular, in this study the video-level classification was based simply on a majority voting of the image labels from each video, mainly due to the small number of operations available (only 53). Given a larger video dataset, one could consider the patient as the bag, along with its GB vascularity label, and the patches extracted from the GB images of the video as the instances. This way, the patient-level GB classification could be improved, without the need of labeling the images extracted from every video of the operation.

As future work, we aim to expand the GBVasc181 dataset by performing more annotations upon the Cholec80 video collection. Moreover, we aim to combine the MIL concept with CNNs at the image-level to improve further the classification performance. In particular, we currently design a 3D CNN architecture that takes as input a sequence of patches extracted from a GB image and outputs the vascularity label of the image. The generation of spatial attention maps that allow visualization of GB wall regions with a variable vascularity is also major topic of interest for future research work.

REFERENCES

[1] T. M. Ward et al., "Surgical data science and artificial intelligence for surgical education," *J. Surg. Oncol.*, vol. 124, no. 2, pp. 221–230, Aug. 2021.
[2] K. Cheng et al., "Artificial intelligence-based automated laparoscopic cholecystectomy surgical phase recognition and analysis," *Surg. Endosc.*, Jul. 2021.
[3] A. Marafioti et al., "CataNet: Predicting remaining cataract surgery duration," *ArXiv*, Jun. 2021.
[4] B. Zhang et al., "Surgical tools detection based on modulated anchoring network in laparoscopic videos," *IEEE Access*, vol. 8, pp. 23748–23758, 2020.
[5] C. Loukas and E. Georgiou, "Smoke detection in endoscopic surgery videos: a first step towards retrieval of semantic events," *Int. J. Med. Robot. Comput. Assist. Surg.*, vol. 11, no. 1, pp. 80–94, Mar. 2015.
[6] I. Funke et al., "Video-based surgical skill assessment using 3D convolutional neural networks," *Int. J. Comput. Assist. Radiol. Surg.*, vol. 14, no. 7, pp. 1217–1225, Jul. 2019.
[7] C. Loukas et al., "Keyframe extraction from laparoscopic videos based on visual saliency detection," *Comput. Methods Programs Biomed.*, vol. 165, pp. 13–23, Oct. 2018.
[8] B. van Amsterdam et al., "Gesture recognition in robotic surgery: A review," *IEEE Trans. Biomed. Eng.*, vol. 68, no. 6, pp. 2021–2035, Jun. 2021.
[9] A. I. Al Abbas et al., "Methodology for developing an educational and research video library in minimally invasive surgery," *J. Surg. Educ.*, vol. 76, no. 3, pp. 745–755, May 2019.
[10] A. P. Twinanda et al., "EndoNet: A deep architecture for recognition tasks on laparoscopic videos," *IEEE Trans. Med. Imaging*, vol. 36, no. 1, pp. 86–97, 2017.
[11] C. Loukas et al., "The contribution of simulation training in enhancing key components of laparoscopic competence," *Am. Surg.*, vol. 77, no. 6, pp. 708–715, 2011.
[12] C. Loukas and N. P. Sgouros, "Multi-instance multi-label learning for surgical image annotation," *Int. J. Med. Robot. Comput. Assist. Surg.*, vol. 16, no. e2058, pp. 1–12, 2020.
[13] S. Madad Zadeh et al., "SurgAI: deep learning for computerized laparoscopic image understanding in gynaecology," *Surg. Endosc.*, vol. 34, no. 12, pp. 5377–5383, 2020.
[14] A. Baghdadi et al., "A computer vision technique for automated assessment of surgical performance using surgeons' console-feed videos," *Int. J. Comput. Assist. Radiol. Surg.*, vol. 14, no. 4, pp. 697–707, 2019.
[15] A. Derathé et al., "Predicting the quality of surgical exposure using spatial and procedural features from laparoscopic videos," *Int. J. Comput. Assist. Radiol. Surg.*, vol. 15, no. 1, pp. 59–67, 2020.
[16] A. Madani et al., "Artificial intelligence for intraoperative guidance: using semantic segmentation to identify surgical anatomy during laparoscopic cholecystectomy," *Ann. Surg.*, 2020 (ahead of print).
[17] P. Beyersdorffer et al., "Detection of adverse events leading to inadvertent injury during laparoscopic cholecystectomy using convolutional neural networks," *Biomed. Tech. Eng.*, 2021 (ahead of print).
[18] C. Loukas et al., "Patch-based classification of gallbladder wall vascularity from laparoscopic images using deep learning," *Int. J. Comput. Assist. Radiol. Surg.*, vol. 16, no. 1, pp. 103–113, 2021.
[19] Y. Iwashita et al., "What are the appropriate indicators of surgical difficulty during laparoscopic cholecystectomy? Results from a Japan-Korea-Taiwan multinational survey," *J. Hepatobiliary. Pancreat. Sci.*, vol. 23, no. 9, pp. 533–547, 2016.
[20] M. Lux and O. Marques, *Visual information retrieval using Java and LIRE*. Morgan & Claypool, 2013.
[21] T. G. Dietterich, et al., "Solving the multiple instance problem with axis-parallel rectangles," *Artif. Intell.*, vol. 89, no. 1–2, pp. 31–71, 1997.
[22] G. Quellec et al., "Multiple-Instance Learning for Medical Image and Video Analysis," *IEEE Rev. Biomed. Eng.*, vol. 10, pp. 213–234, 2017.
[23] J. Amores, "Multiple instance classification: Review, taxonomy and comparative study," *Artif. Intell.*, vol. 201, pp. 81–105, 2013.
[24] O. Maron and T. Lozano-Pérez, "A framework for multiple-instance learning," in *Proceedings of Advances in Neural Information Processing Systems*, 1998, pp. 570–576.
[25] Q. Zhang and S. A. Goldman, "EM-DD: An improved multiple-instance learning technique," in *Advances in Neural Information Processing Systems*, 2001, pp. 1073–1080.
[26] S. Andrews et al., "Support vector machines for multiple-instance learning," in *Advances in Neural Information Processing Systems*, 2003, vol. 15, pp. 1–8.
[27] J. Wang and J.-D. Zucker, "Solving the multiple-instance problem: a lazy learning approach," in *Proceedings of the 17th International Conference on Machine Learning*, 2000, pp. 1119–1126.
[28] Z.-H. Zhu et al., "Multi-instance learning by treating instances as non-I.I.D. samples," in *Proceedings of the 26th International Conference on Machine Learning*, 2009, pp. 1249–1256.
[29] C. M. Bishop, "Illustration: Variational Mixture of Gaussians," in *Pattern Recognition and Machine Learning*, New York: Springer-Verlag New York, Inc., 2006, pp. 474–481.
[30] J. Wang, "MILL: A Multiple Instance Learning Library," 2008. [Online]. Available: http://www.cs.cmu.edu/~juny/MILL/.